\documentclass{article}
\usepackage{ijcai22}

\usepackage{times}
\usepackage{soul}
\usepackage{url}
\usepackage[hidelinks]{hyperref}
\usepackage[utf8]{inputenc}
\usepackage[small]{caption}
\usepackage{graphicx}
\usepackage{amsmath}
\usepackage{amsthm}
\usepackage{booktabs}
\usepackage{algorithm}
\usepackage{algorithmic}
\usepackage{booktabs}
\usepackage{algorithm}
\usepackage{bbding}
\usepackage{graphicx}
\usepackage{subcaption}  
\usepackage{multirow}
\usepackage{amsmath,amsthm}
\usepackage{amsfonts}
\usepackage{mathtools}

\title{Structure-Aware Label Smoothing for Graph Neural Networks}

\author{
Yiwei Wang$^1$
\and
Yujun Cai$^2$\and
Yuxuan Liang$^{1}$\and
Wei Wang$^1$\and
Henghui Ding$^2$\and\\
Muhao Chen$^3$\and
Jing Tang$^4$\And
Bryan Hooi$^1$
\affiliations
$^1$National University of Singapore\\
$^2$Nanyang Technological University\\
$^3$Department of Computer Science, USC\\
$^4$The Hong Kong University of Science and Technology
\emails
wangyw\_seu@foxmail.com, 
\{yujun001, ding0093\}@e.ntu.edu.sg,
yuxliang@outlook.com,
wangwei.cs@gmail.com,
muhaochen@ucla.edu,
jingtang@ust.hk,
bhooi@comp.nus.edu.sg
}

\begin{document}

\maketitle

\begin{abstract}
Representing a label distribution as a one-hot vector is a common practice in training node classification models.
However, the one-hot representation may not adequately reflect the semantic characteristics of a node in different classes, as some nodes may be semantically close to their neighbors in other classes.
It would cause over-confidence since the models are encouraged to assign full probabilities when classifying every node.
While training models with label smoothing can ease this problem to some degree, it still fails to capture the nodes' semantic characteristics implied by the graph structures.
In this work, we propose a novel SALS (\textit{Structure-Aware Label Smoothing}) method as an enhancement component to popular node classification models.
SALS leverages the graph structures to capture the semantic correlations between the connected nodes and generate the structure-aware label distribution to replace the original one-hot label vectors, thus improving the node classification performance without inference costs.
Extensive experiments on seven node classification benchmark datasets reveal the effectiveness of our SALS on improving both transductive and inductive node classification. 
Empirical results show that SALS is superior to the label smoothing method and enhances the node classification models to outperform the baseline methods.   
\end{abstract}

\section{Introduction}
Node classification is a fundamental machine learning task on graphs \cite{wu2019comprehensive}. It supports numerous practical applications, such as learning molecular fingerprints \cite{kearnes2016molecular} and predicting entity properties \cite{schlichtkrull2018modeling}.
Since the introduction of the Graph Convolutional Network \cite{kipf2016semi}, graph neural networks (GNNs) have become the modern tools of choice for node classification.

GNNs devise the `message passing' mechanism, which aggregates the features for every node from its neighbors in the feed-forward process \cite{kipf2016semi}.
`Message passing' offers the inductive bias of reducing the semantic distances between connected nodes \cite{rong2019dropedge,wang2020unifying}.
This inductive bias matches the prior knowledge that the edges naturally imply the connectivity or relatedness between the connected nodes \cite{hartuv2000clustering}, which is the basis of GNNs' superior performance \cite{wang2020unifying}.

On the other hand, how to utilize the graph structural information to generate effective supervision signals remains under-explored.
As far as we know, most of the existing work trains GNNs with the label distribution of one-hot vectors, namely the hard targets \cite{wu2019comprehensive}.
However, recent research has found that using the hard targets to train neural networks tends to cause over-fitting and over-confidence \cite{szegedy2016rethinking,muller2019does}.
To combat these problems, \cite{szegedy2016rethinking} have proposed the label smoothing (LS) technique that applies the uniform noise to the label distribution.
Although LS can prevent the learned models from being over-confident, it cannot describe the characteristics of a node in different classes implied by the graph structures.
For example, if node $i$ in class $y_i$ has a neighbor $j$ in class $y_j$, node $i$ should be more semantically close to $y_j$ than other classes due to the connectivity between nodes $i$ and $j$.
Equally treating all the classes for every node overlooks the graph structures and could limit the GNNs' performance. 

The central idea of this paper is to enrich the supervision of GNNs by incorporating the graph structural information into the label distribution.
We encapsulate this idea in a simple yet effective method, called \textbf{SALS} (\textit{Structure Aware Label Smoothing}), to generate more `faithful' soft targets on node classification.
SALS takes the neighborhood labels as the prior distribution for the label smoothing of every node.
It reflects the semantic characteristics of a target node in different classes implied by the graph structures.
In the semantic space, our SALS reduces the optimum distances from a target node to its neighbors so as to guide the target node's representation toward an appropriate position.
This suits the inductive bias of GNNs and calibrates them by adaptively offering the structure-aware training targets for every node.

SALS is a general method for node classification that offers improvements %in effectiveness 
by regularizing GNN models without extra inference costs.
We evaluate SALS on both transductive and inductive node classification tasks using the Citeseer, Cora, Pubmed \cite{london2014collective}, CoraFull \cite{bojchevski2018deep}, Coauthor-Physics \cite{shchur2018pitfalls}, Flickr \cite{mcauley2012image}, and Reddit \cite{hamilton2017inductive} datasets.
Qualitatively, SALS learns more discriminative node representations (see Fig. \ref{fig:tsne}).
We also observe consistent quantitative improvements measured by test accuracy.
Overall, SALS improves the popular GCN \cite{kipf2016semi}, ResGCN \cite{li2019deepgcns}, GraphSAGE \cite{hamilton2017inductive}, GraphSAINT \cite{zeng2019graphsaint}, and SIGN \cite{frasca2020sign} models by a significant margin, and enhances them to outperform the baseline methods.

\section{Related Work}\label{ref:related}

Graph Neural Networks (GNNs) for node classification have seen a long history of studies, we thus refer readers to \cite{wu2019comprehensive} and \cite{zhou2018graph} for a comprehensive review.
The first work that proposes the convolution operation on graph data is \cite{bruna2013spectral}.
More recently, \cite{kipf2016semi} made breakthrough advancements in the task of node classification.
After \cite{kipf2016semi}, numerous GNN methods have been proposed for better performance on node classification.
There are two main lines of research in this field.
The first one is to propose new GNN architectures to improve the model capacity \cite{velivckovic2017graph,zhang2018gaan,haonan2019graph,zhuang2018dual,qu2019gmnn}.
Another one is to propose new mini-batch training techniques for GNNs to enhance their scalability without the loss of effectiveness \cite{hamilton2017inductive,chiang2019cluster,zeng2019graphsaint}.
A common design among these GNN models is the `message passing' mechanism, which matches the characteristics of graph data that the edges imply the connectivity or relatedness between nodes.

Our work is orthogonal to the above two lines in the sense that it does not alter the GCN architecture, or introduce any new mini-batch technique.
Instead, we propose a new label smoothing method that enriches the supervision of GNNs without introducing extra inference costs.
SALS incorporates the rich graph structures into the supervision of GNNs. 
We find that the favorable characteristics of SALS lead to more accurate predictions.

Label smoothing (LS) is first proposed in image classification tasks as a regularization technique \cite{szegedy2016rethinking}, and has been used in many state-of-the-art models, including computer vision \cite{Zoph_2018_CVPR,zhang2021delving}, and natural language processing \cite{chorowski2016better,Vaswani2017atten}. 
However, the soft targets produced by the LS cannot describe the structural characteristics of nodes implied by the connectivity.
In contrast, we use the graph structural information to guide the label smoothing so as to supervise GNNs with the structure-aware label distribution.
As far as we know, our work is the first to devise the label smoothing method for the graph data.

\begin{figure}[!tb]
	\centering
	\includegraphics[width=1\linewidth]{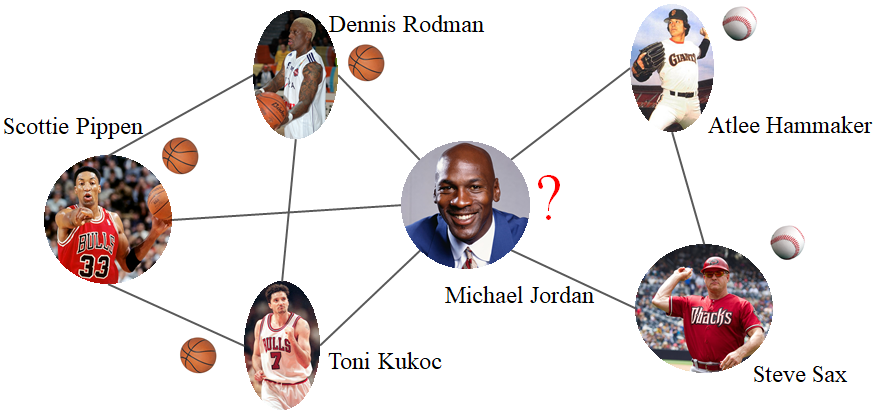}
	\caption{\label{fig:1}\textit{A node's neighborhood is informative on describing its own characteristics.} In a athlete graph, every node is an athelete, and every edge represents a teammate relation. An athlete's neighbors imply its own characteristics. For example, Michael Jordan, a famous basketball athlete, is also professional at baseball, as implied by his neighbors.}
\end{figure}

\section{Methodology}\label{sec:method}
In this section, we introduce the technical details of our proposed structure-aware label smoothing (SALS).
Our SALS incorporates the idea of modeling deterministic data, namely observed class labels, in terms of a set of probability distributions instead of a hard target.
SALS utilizes the rich graph structure information for supervising graph neural networks on node classification, so as to produce more `faithful' training targets reflecting the connectivity or relatedness between the connected nodes.
Consequently, SALS yields the soft target for every node to be adaptive to its neighborhood, which is shown to be crucial to learning effective node embeddings.

\subsection{Preliminaries} \label{sec:3_1}
We %begin by describing the problem of node classification on graphs and introducing notation.
hereby introduce the preliminaries by introducing the node classification problem, graph neural networks, and the original label smoothing technique.

\paragraph{Node Classification} 
Consider a graph $\mathcal{G} = (\mathcal{V}, \mathcal{E})$, where $\mathcal{V}$ is the set of nodes and $\mathcal{E} = \{(i,j)\mid i,j\in \mathcal{V}\}$ is the set of edges.
The goal of node classification is to learn a mapping $\mathcal{M}\colon \mathcal{V} \mapsto \mathbb{P}(\mathcal{Y})$, where $\mathcal{Y}$ is a set of class labels, and $\mathbb{P}(\mathcal{Y})$ is the space of probability distributions over $\mathcal{Y}$.

Node classification has been mainly addressed by Graph Neural Network (GNN) based methods in recent literature.
The edges in a graph imply the connectivity or relatedness between the connected nodes \cite{hartuv2000clustering}.
Based on this prior knowledge, GNNs utilize the connectivity of nodes in graph structures to learn the nodes' representations.
GNNs are a kind of multi-layer neural networks that propagates the nodes' representations across edges between different nodes, which is also known as the `message passing' mechanism.
GNNs stack multiple trainable layers to achieve message passing over edges, where every GNN layer updates a target node $i$'s representations by aggregating the last layer's representations from the target node's neighbors (in the training set) denoted as $\mathcal{N}(i)$.

\begin{figure}[!tb]
	\centering
	\includegraphics[width=0.8\linewidth]{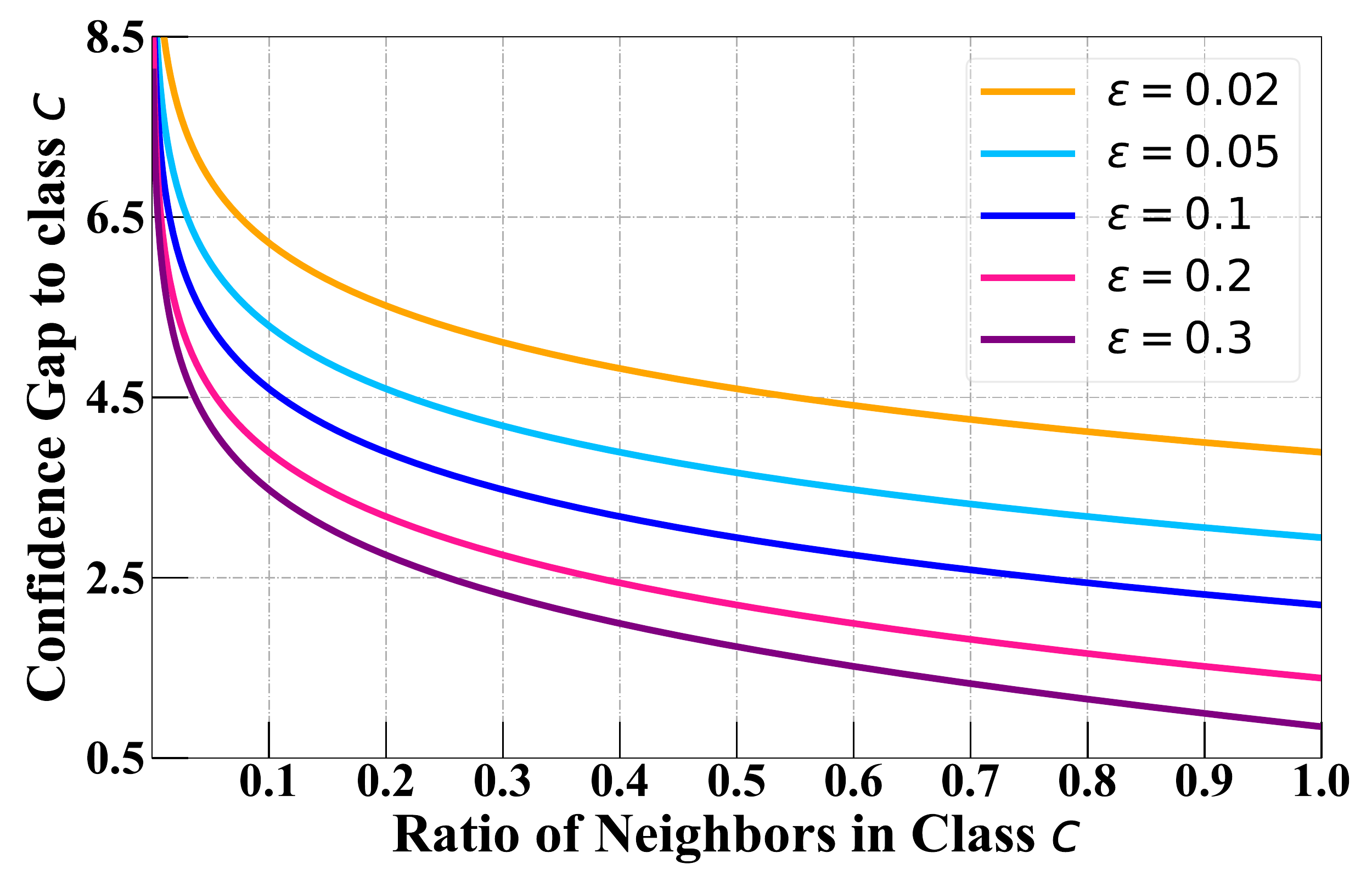}
	\caption{\textit{The gap between the optimum logits on the correct class $y_i$ and another class $c'$ with our SALS (see Eq. \eqref{eq:gap}).} `Ratio of Neighbors' denotes the ratio of the neighbors in class  $c$.
		More neighbors in class $c$ corresponds to a smaller gap from the correct class to class $c$ with our SALS.
		\label{fig:gap}}
\end{figure}

Existing work finds that `message passing' reduces the distance in the embedding space between the connected nodes \cite{rong2019dropedge,wang2020unifying}.
This property matches the characteristics of the graph data that every node is correlated to its neighbors. 
It acts as a useful inductive bias that supports GNNs' state-of-the-art performance.

\paragraph{Label Smoothing} 
The classical training is based on the deterministic label, which is (explicitly or implicitly) treated as a one-hot vector, namely the hard targets.
The recent study \cite{muller2019does} finds that the hard targets are too extreme to express the normally realistic assumption of a non-deterministic dependency between the features and labels.
Some instances are semantically close to other instances in the classes different from themselves \cite{zhang2021delving}.
Instead of using hard labels for training, label smoothing (LS) replaces the surrogate distribution $q$ with a less extreme surrogate 
\begin{equation}
q^{\mathrm{LS}} = (1 - \epsilon)q + \epsilon u_c
\end{equation}
as a soft target for the learner, where $u_c \in \mathbb{P}(\mathcal{Y})$ is a uniform distribution among all the classes in $\mathcal{Y}$ and $\epsilon \in (0, 1]$ is a smoothing factor.
LS can be seen as an attempt at presenting the training information in a more `faithful' way: a smoothed target probability $q^{\mathrm{LS}}$ is arguably more realistic than a degenerated hard distribution $q$ assigning the full probability mass to a single class label.

LS obviously leads to less extreme predictions than the hard targets.
However, the adjusted distribution $q^{\mathrm{LS}}$ assigns the uniform noise to all the classes, which may not reflect the intrinsic characteristics of an instance to be classified.

\subsection{Structure-Aware Label Smoothing}
For training GNNs, most of existing work minimizes the expected value of the cross-entropy loss between the true targets $q(c|i)$ and the network's prediction $p(c|i)$ in 
\begin{equation}
H(q, p) = -\sum_{c = 1}^{C}q(c|i) \log p(c|i).
\end{equation}
Minimizing this loss is equivalent to maximizing the expected log-likelihood of the label that is selected according to the distribution $q(c|i)$, of which the gradient is bounded between $-1$ and $1$ \cite{szegedy2016rethinking}.

Consider the case of the hard target, where $q(y_i|i) = 1$ and $q(c|i) = 0$ for $\forall c \neq y_i$, training GNNs with hard targets has several limitations.
First, it overlooks the correlations between connected nodes.
The edges in a graph naturally imply the connectivity or relatedness between connected nodes \cite{hartuv2000clustering}.
Second, using hard targets to supervise GNNs does not match their inductive bias of reducing the semantic distances between connected nodes \cite{wang2020unifying}.
To classify node $i$, the `message passing' mechanism of GNNs aggregates the features from $i$'s neighbors, while its neighbor $j \in \mathcal{N}(i)$ can belong to any other classes than $y_i$.
Enforcing GNNs to assign full probabilities as prediction for connected nodes in different classes ignores their structural relationship, and is hard to generalize to the unseen nodes.

To address the above issues, intuitively, we aim to find a label distribution that reveals the semantic characteristics of a node in each class implied by the graph structures.
Consider an example in Fig. \ref{fig:1}.
In a social network of athletes where the edges represent the `teammate' relations, the node `Michael Jordan' is a famous professional basketball athlete, and he is meanwhile an expert in baseball, which is implied by his neighbors of professional baseball athletes \cite{mathur1997wealth}. 
Here, `basketball' and `baseball' should be assigned a higher probability mass than the other sports, e.g., `soccer', at which the node `Michael Jordan' is not a professional.
It would be beneficial to consider these genuine `preferences' of a node on different classes implied by the graph structures for label smoothing.

We propose a method called \textbf{SALS} (\textit{Structure-aware Label Smoothing}) to mine the intrinsic semantic characteristics of every node from rich graph structures for node classification, and produce structure-aware soft targets for training GNNs.
Specifically, we use the labels of a node's neighbors to mine its characteristics beyond its own hard target, and use the structural information to produce the soft targets that are adaptive to the graph structures.
SALS regularizes GNNs for better generalization and makes node representation more adaptable by leveraging the rich graph structural knowledge for supervision.
For node $i$ with ground-truth labels $y_i$, consider the distributions of its neighbors' labels $\left\{y_j, j\in \mathcal{N}(i)\right\}$.
First, we analyze the ratio of neighbors (in the training set) of nodes $i$ in different classes.
In particular, let the ratio of node $i$'s neighbors in class $y$ be
\begin{equation} \label{eq:ratio}
r_{c}(i) \coloneqq \frac{\sum_{j \in \mathcal{N}(i)}{1[y_j = c]}}{|\mathcal{N}(i)|}.
\end{equation}
We define the neighborhood label of node $i$, which is the average label of $i$'s neighbors in the training set, to be
\begin{equation}
\eta(i) \coloneqq \frac{1}{|\mathcal{N}(i)|}\sum_{j\in \mathcal{N}(i)} \delta_{c, y_j} = \sum_{y\in \mathcal{Y}} r_{y}(i) \delta_{c,y},
\end{equation}
where $\delta_{c,y_j}$ is Dirac delta, which equals 1 for $c = y_j$ and 0 otherwise.
We replace the label distribution of $q(c|i) = \delta_{c,y_i}$ with
\begin{align}\label{eq:SLS}
q^{\mathrm{SALS}}(c|i) = (1 - \epsilon)\delta_{c,y_i} + \epsilon \Big(\gamma \eta(i) + (1 - \gamma) u_c\Big),
\end{align}
which is a mixture of the original hard target $q(c|i)$ and its neighbors' 
%bh2: neighbors'
label distributions, with weights $1-\epsilon$ and $\epsilon$, respectively.
$\epsilon$ is a hyper-parameter for weighting the label smoothing.
$\gamma$ is the `factor' balancing our neighborhood label and the pre-defined uniform smoothing from LS \cite{muller2019does}.
Our SALS introduced in Eq. \eqref{eq:SLS} can be seen as the distribution of node $i$'s label obtained as follows: first, set it to the ground-truth label $c = y_i$; then, with probability $\epsilon$, replace $y_i$ with a sample drawn from the distribution $\gamma \eta(i) + (1 - \gamma) u_c$, which is the prior distribution over labels of node $i$'s neighborhood.

\begin{figure}[!tb]
	\centering
	\includegraphics[width=1\linewidth]{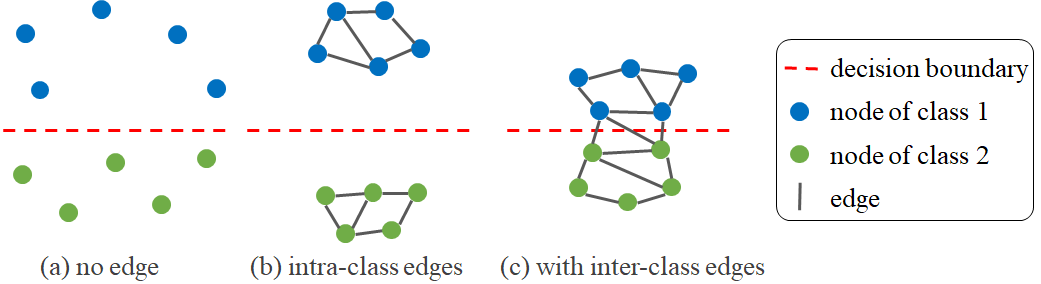}
	\caption{\textit{The message passing of GNNs reduce the embedding distances between the connected nodes.}  We compare the node embeddings retrieved by GNNs to a physical equilibrium mode, where edges connecting nodes serve as the rubber bands that expose explicit constraints to pull the connected nodes together. \label{fig:sys}}
\end{figure}

Note that SALS achieves desired goals of utilizing the graph structural information for supervising GNNs.
It encourages the target node's penultimate layer representation to be close to the template of the correct class, and meanwhile closer to its neighbors' class templates than to other incorrect classes.
This matches the prior knowledge that the edges between nodes represent connectivity or relatedness. 
Meanwhile, it matches the inductive bias from `message passing', that the nodes' features are aggregated along the edges, by which GNNs tend to give similar predictions to the connected nodes \cite{wang2020unifying}.

When training GNNs with our SALS, the gap between the optimum logits on the correct class $y_i$ and another class $c'$, which minimizes the cross entropy loss, is
\begin{align}\label{eq:gap}
\begin{split}
&(\mathbf{x}_i^T\mathbf{w}_{y_i})^\star - (\mathbf{x}_i^T\mathbf{w}_{c'})^\star \\
= &\log\left(\frac{1 - \epsilon + \epsilon\gamma r_c(i) + \epsilon(1 - \gamma) / |\mathcal{Y}|}{\epsilon\gamma r_c(i) + \epsilon(1 - \gamma) / |\mathcal{Y}|}\right),
\end{split}
\end{align}
where $\mathbf{x}_i$ is the final layer representation of node $i$ and $\mathbf{w}_c$ is the template of class $c$.
We visualize this gap in Fig. \ref{fig:gap} with $\gamma = 1$.
With more neighbors in class $c'$, this gap shrinks, which validates that our SALS reflects the influence of the neighborhood in class $c'$.
When $\epsilon \rightarrow 0$, SALS degrades to the hard targets.
In this case, the gap in Eq. \eqref{eq:gap} is $\infty$, which is unachievable given finite values of logits.
In contrast, our SALS makes the target gaps be finite values, which are more `practical' training targets for GNNs, so as to reduce the over-confidence problems.
Overall, our SALS enriches the supervision signals with graph structural knowledge, which regularizes GNNs to adapt to complex graph topology.

\begin{figure}[!tb]
	\centering
	\includegraphics[width=0.5\textwidth]{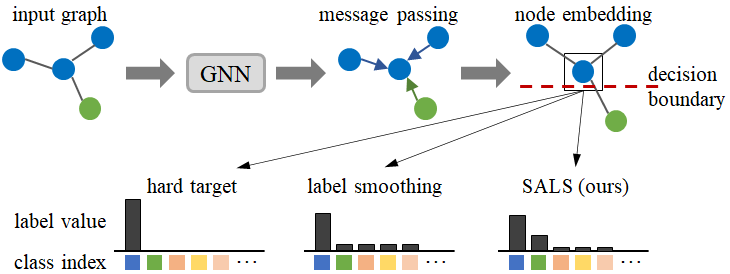}
	\caption{\textit{Our SALS produces the soft targets for every node adaptive to graph structures.} The structure-aware soft targets produced by SALS match the inductive bias of GNNs' message passing that reduces the semantic distances between the connected nodes. \label{fig:com}}
\end{figure}

\subsection{Discussion}\label{sec:3.2}

In this section, we analyze how our SALS method influences the supervision of node classification.
An interpretation of SALS can be obtained by considering the cross entropy:
\begin{align}\label{eq:30}
&H\left(q^{\mathrm{SALS}}(c|i), p(c|i)\right) \nonumber\\
=& (1 - \epsilon)H\left(q(c|i), p(c|i)\right) + \\ \nonumber
&\epsilon \Big(\frac{\gamma}{|\mathcal{N}(i)|}\sum_{j \in \mathcal{N}(i)} H\left(q(c|j), p(c|i)\right) + (1 - \gamma)H(u, p) \Big)
\end{align}
Thus, SALS is to add a group of losses $\left\{H\left(q(c|j), p(c|i)\right), j \in \mathcal{N}(i)\right\}$, which penalizes the deviation of predicted label distribution $p(c|i)$ from the label distributions of neighbors.
Note that this deviation could be equivalently captured by the KL divergence, which is a measure of how dissimilar the predicted distribution $p(c|v)$ is to its neighbors' label distributions.
This is in accordance with the characteristics of the graph data.
In a graph, edges represent the natural connectivity or relatedness between nodes \cite{hartuv2000clustering}.
Motivated by this, we use the loss in Eq. \eqref{eq:30} to encourage the nodes to have the final-layer representations that are closer to the class templates of their neighbors.

To intuitively understand the role of our SALS, we draw an analogy with a physical equilibrium model as shown in Fig. \ref{fig:sys}.
Each node is seen as a particle, while the supervised classification signals act as the implicit force pulling the nodes away from the decision boundary.
Without edges (Fig. \ref{fig:1}(a)), nodes are loosely placed in the embedding space.
In contrast, with message passing, edges act as the rubber bands and expose explicit constraints on the representations of connected nodes.
In an ideal case where edges only connect nodes with the same label, the message passing will pull nodes within the same class together, which greatly benefits classification (Fig. \ref{fig:1}(b)).

The connected nodes may have different class labels from each other.
These edges pull the nodes of low $r_{y_i}$ (see Eq. \eqref{eq:ratio}) towards the decision boundaries (see Fig. \ref{fig:1}(c)).
If training with the hard targets, the nodes of lower $r_{y_i}$ have larger losses and gradient magnitudes (on the final layer), since these nodes are closer to the decision boundary than the nodes of higher $r_{y_i}$.
These gradients of higher magnitudes influence the updating of GNNs' learnable weights more heavily during training.
Enforcing the nodes of low $r_{y_i}(i)$ to be far away from the decision boundary can lead to the over-confidence problem, and is hard to generalize to unseen nodes.
This is meanwhile conflicted with the graphs' characteristics that the connected nodes are semantically close to each other.
On the other hand, since the gradients on the nodes of lower $r_{y_i}(i)$ dominate the gradients for updating GNNs, the optimizer takes relatively lower efforts to form the discriminative representations for the nodes of higher $r_{y_i}(i)$.

In contrast, with our SALS, the soft targets (see Eq. \eqref{eq:SLS}) reflect the influence of neighbors in different classes.
As a result, the gradient magnitudes of lower $r_{y_i}(i)$ are balanced with the nodes of higher $r_{y_i}(i)$.
Moreover, if the number of nodes in class $c$ increases, the influence from the class $c$ improves, and our SALS's soft can reflect this change with higher target probability on class $c$.
These structure-aware soft targets align well with the prior knowledge of the graph structures.
SALS matches the inductive bias of graph neural networks and can generalize to the unseen nodes in a better calibrated manner with more appropriate confidence targets (see Fig. \ref{fig:cali}).
Overall, our soft targets correspond to the appropriate positions in the embedding space, which are adaptive to graph structures, and meanwhile maintain the correctness of classification.

\section{Experiments}

In this section, we evaluate the effectiveness of our SALS by applying it to various GNN architectures. 
We report the experimental results under both the transductive and inductive settings.
In the transductive setting, the training phase has access to the features of all nodes but only the labels of nodes in the training set.
In the inductive setting, neither the features nor labels of nodes in the validation/testing set are available during training.
In addition, we visualize the learned representations of GNNs with SALS compared with those from the original GNNs without SALS. 
Last but not least, we conduct ablation studies to show the influence of SALS, as well as the sensitivity with respect to the hyper-parameters.

We use seven benchmark datasets: Cora, Citeseer, Pubmed \cite{london2014collective}, CoraFull \cite{bojchevski2018deep}, Coauthor-Physics (short as Coauthor-Phy) \cite{shchur2018pitfalls}, Flickr \cite{mcauley2012image}, and Reddit \cite{zeng2019graphsaint} for evaluation. 
The former three are citation networks, where each node is a document and each edge is a citation record. 
CoraFull is the larger version of the well-known citation network Cora dataset, where nodes represent publications and edges represents their citations, and the nodes are labeled based on the research topics.
In Flickr, each node represents one image. 
An edge is built between two images if they share some common properties (e.g., same geographic location, same gallery, etc.). 
Reddit is collected from an online discussion forum where users comment in different topical communities. Two posts (nodes) are connected if some users comment on both posts.
Each of them contains an unweighted adjacency matrix and bag-of-words features.
The statistics of these datasets are summarized in Table \ref{tab:data}. 

For the hyper-parameters of baselines,
%%ww: benchmarks --> baselines.  "benchmark datasets"
e.g., the number of hidden units, the optimizer, the learning rate, etc., we set them as suggested by their authors. 
For the hyper-parameters of our SALS, we set $\epsilon = 0.4, \gamma = 0.8$ for SALS by default.
Note that this setting holds for all of our experiments unless otherwise specified. 

\begin{table}[tb!]
	\centering
	\caption{Statistics of the datasets for node classification.}
	\label{tab:data}
	\begin{tabular}{@{}l| c c c c@{}}
		\toprule
		\textbf{Dataset}
		& $\#$\textbf{Nodes}
		& $\#$\textbf{Edges}
		& $\#$\textbf{Classes} \\ \midrule\midrule
		\texttt{Cora} & 2,708 & 5,429 & 7 \\
		\texttt{Citeseer} & 3,327 & 4,732 & 6 \\
		\texttt{Pubmed} & 19,717 & 44,338 & 3\\
		\texttt{CoraFull} & 19,793 & 65,311 & 70 \\
		\texttt{Coauthor-Phy} & 34,493 & 247,962 & 5 \\
		\texttt{Flickr} & 89,250 & 899,756 & 7 \\
		\texttt{Reddit} & 232,965 & 11,606,919 & 41 \\
		\bottomrule
	\end{tabular}
\end{table}

\begin{table*}[tb!]
	\centering
	\caption{Test Accuracy (\%) of transductive node classification. We conduct 100 trials with random weight initialization. The mean and standard derivations are reported.}
	\label{tab:trans_exp}
	\begin{tabular*}{\textwidth}{@{\extracolsep{\fill}} l c c c c c c @{}}
		%\resizebox{\textwidth}{!}{
		%    \begin{tabular}{@{}l c c c c c c @{}}
		\toprule
		\textbf{Method}
		& \texttt{Citeseer}
		& \texttt{Cora}
		& \texttt{Pubmed} 
		& \texttt{CoraFull}
		& \texttt{Coauthor-Phy}
		\\ \midrule\midrule
		GCN \cite{kipf2016semi} & 77.1 $\pm$ 1.4 & 88.3 $\pm$ 0.8 & 86.4 $\pm$ 1.1 & 64.5 $\pm$ 1.3 & 96.0 $\pm$ 0.5\\ 
		GAT \cite{velivckovic2017graph}& 76.3 $\pm$ 0.8 & 87.6 $\pm$ 0.5 & 85.7 $\pm$ 0.7 & 65.3 $\pm$ 0.9 & 96.2 $\pm$ 0.8\\
		JKNet \cite{xu2018representation} & 78.1 $\pm$ 0.9 & 89.1 $\pm$ 1.2 & 86.9 $\pm$ 1.3 & 65.0 $\pm$ 1.1 & 95.1 $\pm$ 0.5\\
		LGCN \cite{gao2018large} & 77.5 $\pm$ 1.1 & 89.0 $\pm$ 1.2 & 86.5 $\pm$ 0.6 & 64.6 $\pm$ 1.0 & 95.4 $\pm$ 0.6\\
		GMNN \cite{qu2019gmnn} & 77.4 $\pm$ 1.5 & 88.7 $\pm$ 0.8 & 86.7 $\pm$ 1.0 & 64.9 $\pm$ 1.1 & 95.6 $\pm$ 0.9\\
		ResGCN \cite{li2019deepgcns} & 77.9 $\pm$ 0.8 & 88.1 $\pm$ 0.6 & 87.1 $\pm$ 1.2& 64.8 $\pm$ 1.2 & 95.3 $\pm$ 1.0\\
		\midrule
		GCN + LS \cite{szegedy2016rethinking} & 77.5 $\pm$ 1.3 & 88.5 $\pm$ 0.6 & 86.7 $\pm$ 0.6 & 64.9 $\pm$ 1.1 & 96.2 $\pm$ 0.5\\
		ResGCN + LS \cite{szegedy2016rethinking} & 78.1 $\pm$ 0.9 & 88.3 $\pm$ 0.7 & 87.2 $\pm$ 0.9 & 65.2 $\pm$ 1.3 & 95.6 $\pm$ 1.0
		\\
		\midrule
		GCN + OLS \cite{zhang2021delving} & 77.7 $\pm$ 1.1 & 88.6 $\pm$ 1.0 & 86.8 $\pm$ 0.7 & 65.0 $\pm$ 1.2 & 96.1 $\pm$ 0.7\\
		ResGCN + OLS \cite{zhang2021delving} & 78.2 $\pm$ 0.9 & 88.3 $\pm$ 0.6 & 87.3 $\pm$ 0.8 & 65.1 $\pm$ 1.1 & 95.4 $\pm$ 0.9
		\\
		\midrule
		GCN + SALS (Ours) & 77.8 $\pm$ 1.0 & \textbf{88.9 $\pm$ 0.7} & 87.6 $\pm$ 0.6 & 66.2 $\pm$ 1.0 & \textbf{96.3 $\pm$ 0.4}\\
		ResGCN + SALS (Ours) & \textbf{78.4 $\pm$ 0.9} & 88.7 $\pm$ 0.7 & \textbf{87.9 $\pm$ 0.8} & \textbf{66.5 $\pm$ 1.1} & 95.8 $\pm$ 0.9
		\\
		\bottomrule
		%\end{tabular}
	\end{tabular*}
	%}
\end{table*}

\subsection{Transductive Node Classification}
In the transductive settings, we take the popular GNN models of GCN \cite{kipf2016semi}, GAT \cite{velivckovic2017graph}, JKNet \cite{xu2018representation}, LGCN \cite{gao2018large},  GMNN \cite{qu2019gmnn}, ResGCN \cite{li2019deepgcns}, the original label smoothing (LS), and the recently proposed Online Label Smoothing (OLS) for image classification \cite{zhang2021delving} as the baselines for comparison.
We follow the prior studies \cite{xu2018representation} to split nodes in each graph into 60\%, 20\%, 20\% for training, validation, and testing for a fair comparison.
We make 10 random splits and conduct the experiments for 100 trials with random weight initialization for each split.

We vary the number of layers from 1 to 10 for each model and choose the best performing number with respect to the validation set.
The results are reported in Table \ref{tab:trans_exp}.
We observe that SALS improves the test accuracy of GCN by 0.9\% on Citeseer, 0.7\% on Cora, 1.4\% on Pubmed, 2.6\% on CoraFull, 0.3\% on Coauthor-Phy, and improves ResGCN by 0.6\% on Citeseer, 0.7\% on Cora, 0.9\% on Pubmed, 2.7\% on CoraFull, 0.5\% on Coauthor-Phy respectively.
As a result, SALS regularizes GCN and ResGCN to outperform all the baseline methods.

Taking a closer look, we find that given the same GNN model, SALS consistently produces larger improvements than LS and OLS.
The advantages come from our structure-aware soft targets that utilize the structural information on label smoothing, while neither LS nor OLS utilizes the connectivity characteristics of the graph data for supervising GNNs.
SALS regularizes GNNs to adapt to the complex topology for every node at different positions of the graph.

\begin{table}[tb!]
	\centering
	\caption{Test accuracy (\%) of inductive node classification. We report mean and standard derivations of 100 trials with random weight initialization. We implement LS, OLS, and our SALS with GraphSAGE, GraphSAINT, and SIGN.}
	\label{tab:induc_exp}
	\begin{tabular}{@{} l  c c @{}}
		\toprule
		\textbf{Method}
		& \texttt{Flickr}
		& \texttt{Reddit} \\ \midrule\midrule
		GraphSAGE & 50.1  $\pm$  1.1 & 95.3  $\pm$  0.1\\
		GraphSAGE + LS & 50.2 $\pm$ 0.7 & 95.5 $\pm$ 0.1\\
		GraphSAGE + OLS & 50.5 $\pm$ 0.8 & 95.6 $\pm$ 0.1\\ \midrule
		GraphSAGE + SALS (Ours) & \textbf{51.1 $\pm$ 0.9} & \textbf{96.0 $\pm$ 0.1}\\ \midrule\midrule
		GraphSAINT & 51.1 $\pm$ 0.2 & 96.6 $\pm$ 0.1 \\ 
		GraphSAINT + LS & 51.4 $\pm$ 0.3 & 96.7 $\pm$ 0.1\\ 
		GraphSAINT + OLS & 51.2 $\pm$ 0.2 & 96.6 $\pm$ 0.1\\ \midrule
		GraphSAINT + SALS (Ours) & \textbf{51.9 $\pm$ 0.2} & \textbf{96.9 $\pm$ 0.1}\\
		\midrule\midrule
		SIGN & 51.4 $\pm$ 0.1 & 96.8 $\pm$ 0.0 \\ 
		SIGN + LS & 51.5 $\pm$ 0.1 & 96.8 $\pm$ 0.1\\ 
		SIGN + OLS & 51.7 $\pm$ 0.1 & 96.9 $\pm$ 0.1\\ \midrule
		SIGN + SALS (Ours) & \textbf{52.2 $\pm$ 0.1} & \textbf{97.1 $\pm$ 0.1}\\
		\bottomrule
		%    \end{tabular}
		%    }
	\end{tabular}
\end{table}

\subsection{Inductive Node Classification}
In the inductive settings, we follow the existing work to use the datasets Flickr, Reddit with the fixed partition for evaluation \cite{zeng2019graphsaint,frasca2020sign}.
These datasets can be too large to be handled well by the full-batch training.
Hence, existing work devises the advanced scalable GraphSAGE \cite{hamilton2017inductive} and GraphSAINT \cite{zeng2019graphsaint}, and SIGN \cite{frasca2020sign} models to deal with the large-scale graphs. 
We take them as the baselines for comparison.
We implement SALS with GraphSAGE, GraphSAINT, and SIGN to study whether SALS can improve the performance of GNNs under the inductive setting.

We vary the number of layers of each method from 1 to 10 for each model and choose the best performing model with respect to the validation set. 
The results are reported in Table \ref{tab:induc_exp}.
We observe that SALS improves the test accuracy of GraphSAGE by 2.0\% on Flickr, 0.7\% on Reddit, GraphSAINT by 1.6\% on Flickr, 0.3\% on Reddit, and SIGN by 1.6\%, 0.3\% on Reddit respectively.
As a result, SALS enhances them to outperform the baseline methods. 

Overall, the results above demonstrate the effectiveness of SALS to improve a wide range of GNN models in terms of both transductive and inductive node classification, and consistently outperforms LS and OLS.
%Our SALS regularizes GNNs to offer better effectiveness without extra inference costs.

\begin{figure}[!tb]
	\centering
	\begin{subfigure}[t]{0.233\textwidth}
		\includegraphics[width=\textwidth]{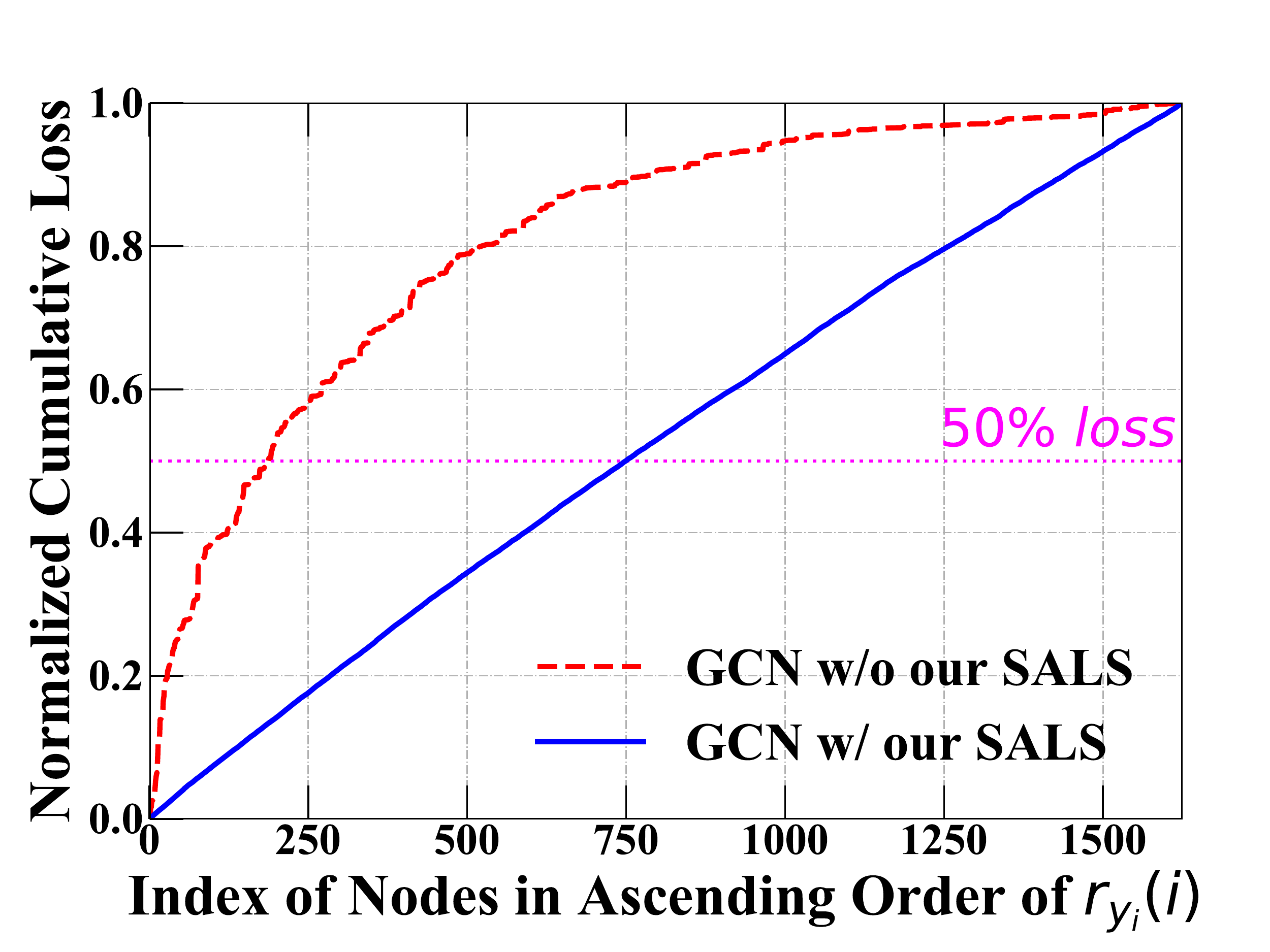}
		\caption{Accumulative Losses}
	\end{subfigure}    \hfill
	\begin{subfigure}[t]{0.233\textwidth}
		\includegraphics[width=\textwidth]{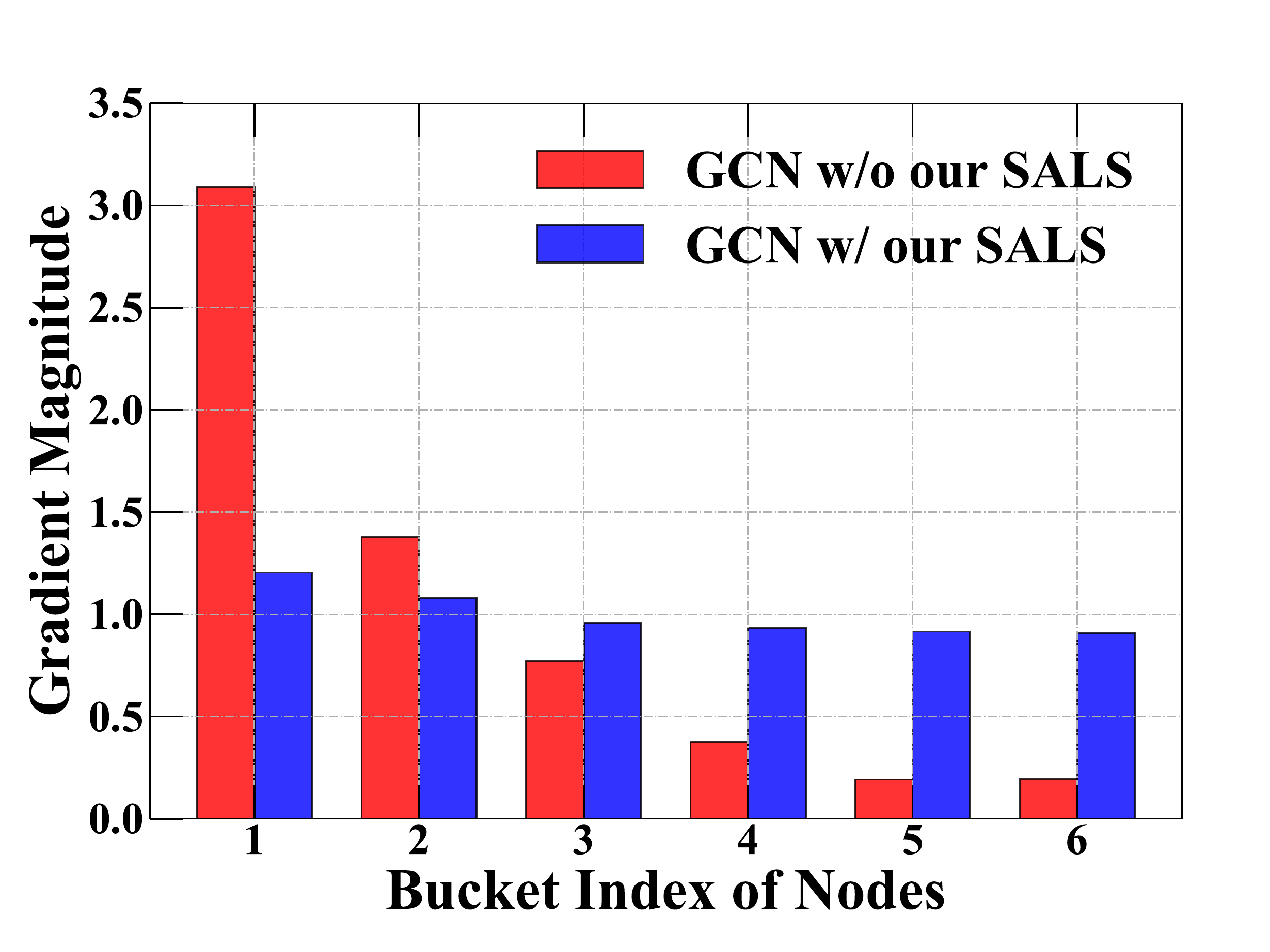}
		\caption{Gradient Magnitudes}
	\end{subfigure}
	\caption{The losses and gradient norms on nodes of different $r_{y_i}(i)$. (a) The normalized accumulative classification loss (y-axis) versus the index of nodes in Citeseer in the ascending order of $r_{y_i}(i)$ (x-axis). (b) The average $\ell_2-$norm of gradients on the nodes in a bucket (y-axis) versus the index of node buckets in Pubmed in the ascending order of $r_{y_i}(i)$ (x-axis). Without our SALS, the nodes of lower $r_{y_i}(i)$ dominate the training losses and the gradients magnitudes, while our SALS balances the losses and gradients.
		\label{fig:dist}}
\end{figure}

\subsection{Visualization and Ablation Study}

We conduct a number of visualization and ablation studies to analyze our SALS. 
First, we investigate the distribution of the classification losses and the final-layer gradients on nodes of different $r_{y_i}(i)$.
In Fig. \ref{fig:dist}, we sort the nodes in the training set by the ascending order of $r_{y_i}(i)$ as defined in Eq. \eqref{eq:ratio}, i.e., the ratio of neighbors that are in the same class as the target node.
We train a 3-layer GCN with and without SALS on different datasets and record the node-wise classification losses and gradients after convergence.
In Fig \ref{fig:dist}(a), we visualize the normalized accumulative losses for the sorted nodes in the Citeseer dataset, where we linearly normalize the losses so that the summation of losses on all the nodes is 1.
In addition, in Fig \ref{fig:dist}(b), we split all the training nodes in the Pubmed dataset into six even buckets of nodes in the ascending order of $r_{y_i}(i)$ and present the average $\ell_2$-norm of gradients in each bucket at the $y$-axis. 

\begin{figure}[!tb]
	\centering
	\begin{subfigure}[t]{0.23\textwidth}
		\includegraphics[width=\textwidth]{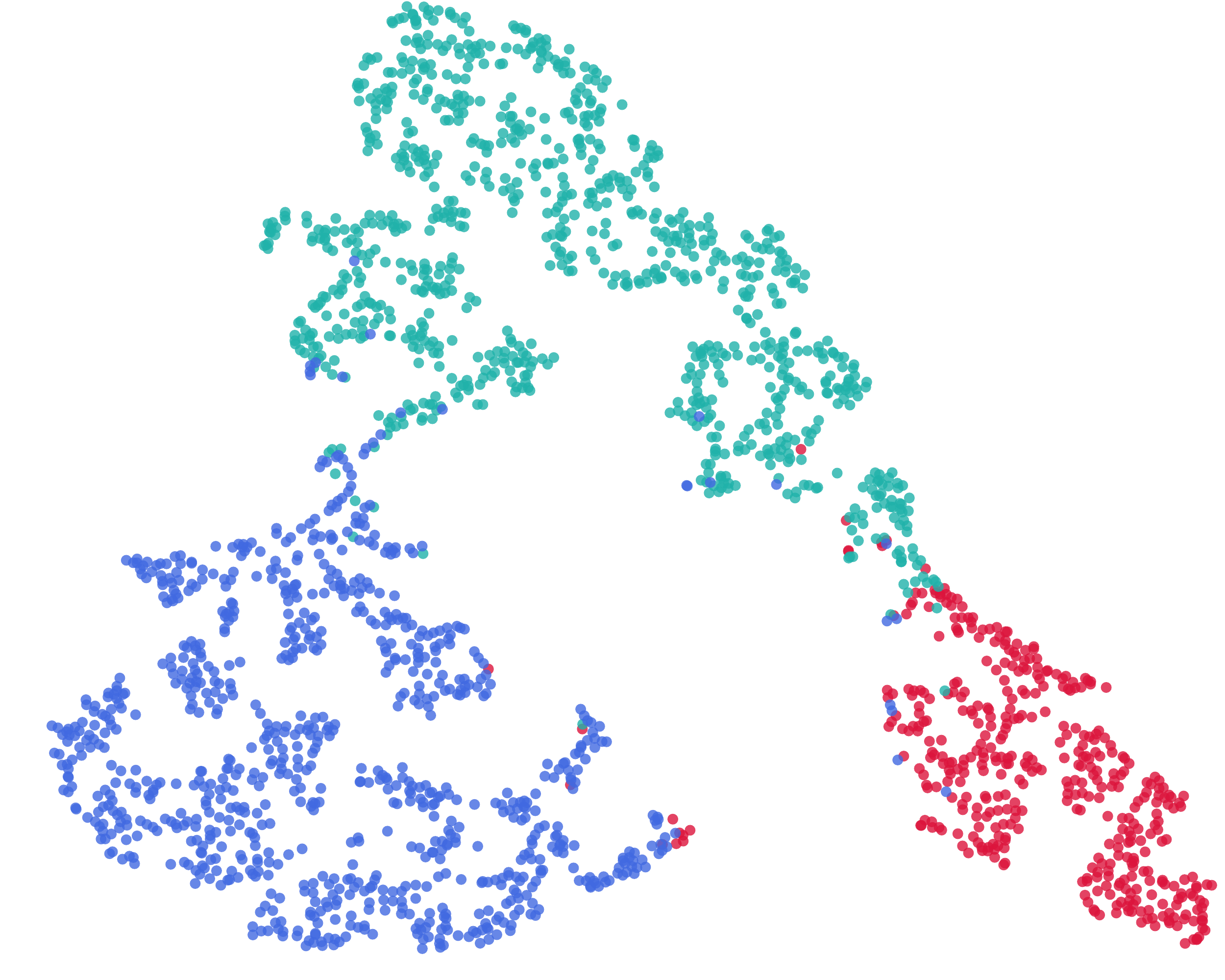}
		\caption{GCN}
	\end{subfigure}    \hfill
	\begin{subfigure}[t]{0.23\textwidth}
		\includegraphics[width=\textwidth]{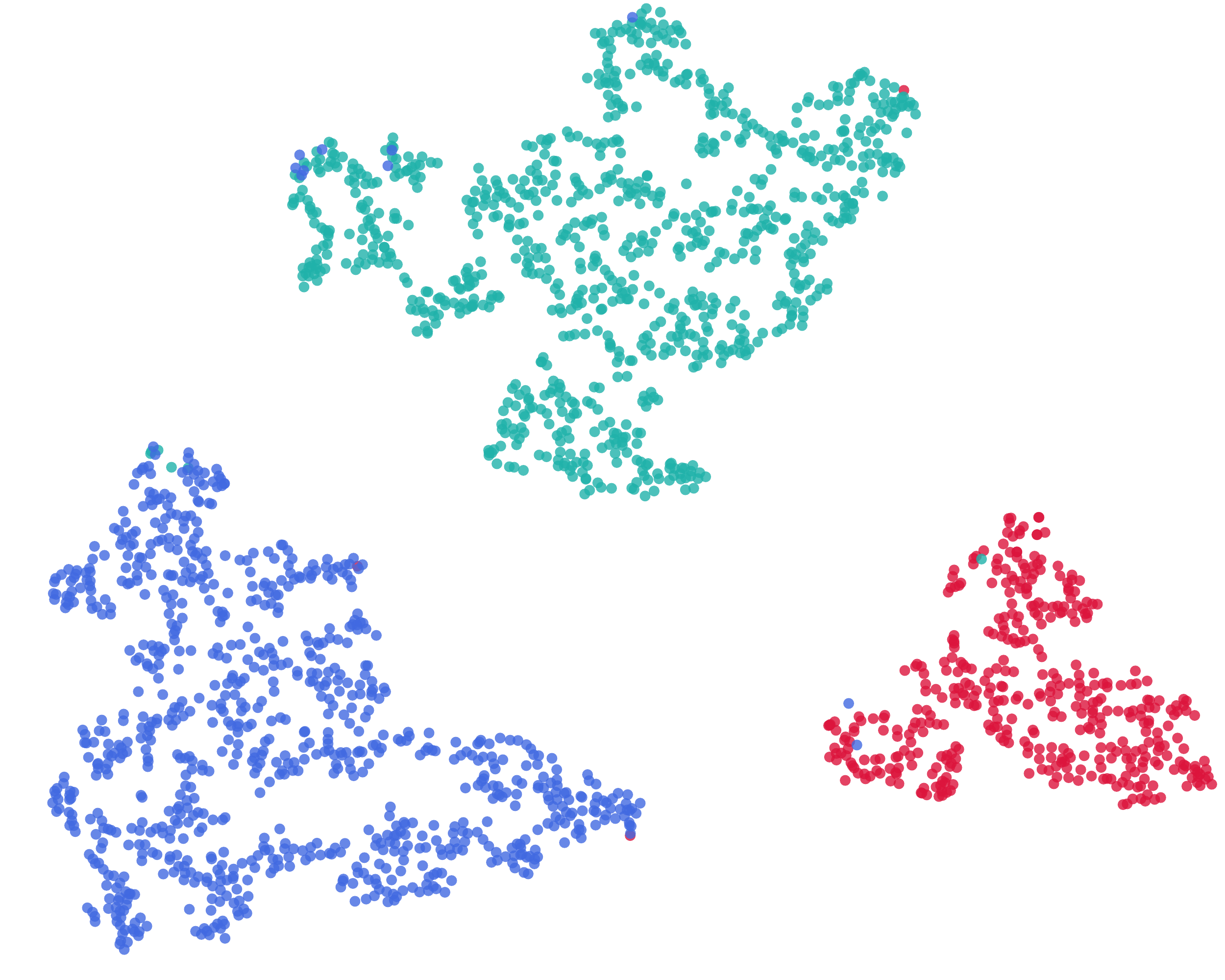}
		\caption{GCN + SALS (Ours)}
	\end{subfigure}
	\caption{The learned representations of nodes in the Pubmed dataset (visualized by t-SNE). Colors denote the ground-truth class labels. The node representations given by GCN with our SALS are more discriminative than those given by the original GCN. \label{fig:tsne}}
\end{figure}

Without our SALS, the nodes of smaller $r_{y_i}(i)$ contributes more to the classification loss and the gradient magnitudes, since GNNs reduce the semantic distances between the connected nodes but the hard targets do not reflect this structural information.
In contrast, with our SALS, the losses on the nodes of different $r_{y_i}(i)$ are better balanced thanks to the structure-aware soft targets.
In this way, the nodes with lower $r_{y_i}(i)$ no longer dominate the weight updating of GNNs' training, so that GNNs can be trained to form more discriminative representations for the nodes of higher $r_{y_i}(i)$.

Fig. \ref{fig:tsne} presents the learned representations obtained by a 3-layer GCN trained with hard targets and one with SALS, where we visualize the representations of the half of nodes in the Pubmed's training set of higher $r_{y_i}(i)$.
It is shown that the hidden layers supported by SALS learn more discriminative features, thanks to the regularization given by SALS.
This meets our analysis in the Section \ref{sec:3.2} and observations from Fig. \ref{fig:dist}.
The structure-aware soft targets produced by our SALS supervised GNNs to have more balanced gradient magnitudes on the nodes of different $r_{y_i}(i)$, which enables GNNs to take more efforts to learn more discriminative node representations than the original hard targets.
These highly discriminative features potentially help produce better class predictions than less discriminative features.

\begin{figure}[!tb]
	\centering
	\includegraphics[width=0.75\linewidth]{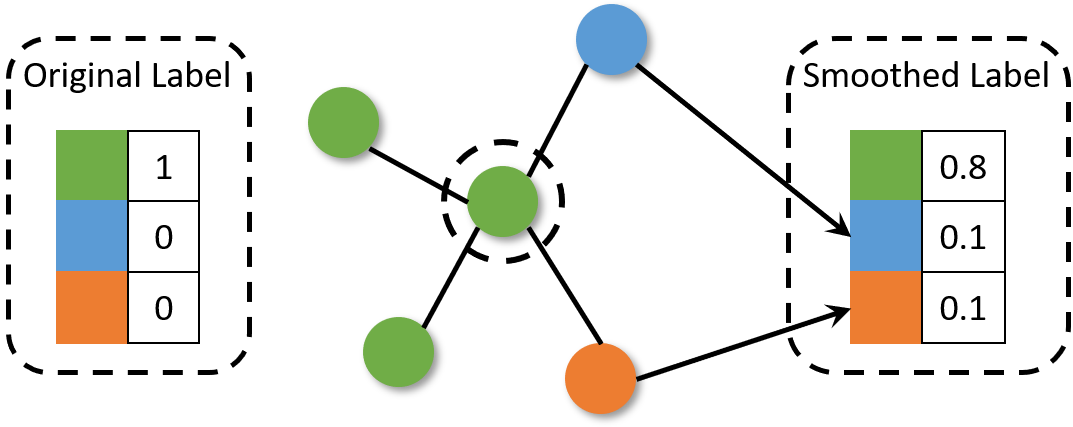}
	\caption{Reliability diagrams of a 3-layer GCN, and it with LS, SALS on the Cora dataset. \label{fig:cali}}
\end{figure}

\begin{figure}[!tb]
	\centering
	\begin{subfigure}[t]{0.48\linewidth}
		\includegraphics[width=\textwidth]{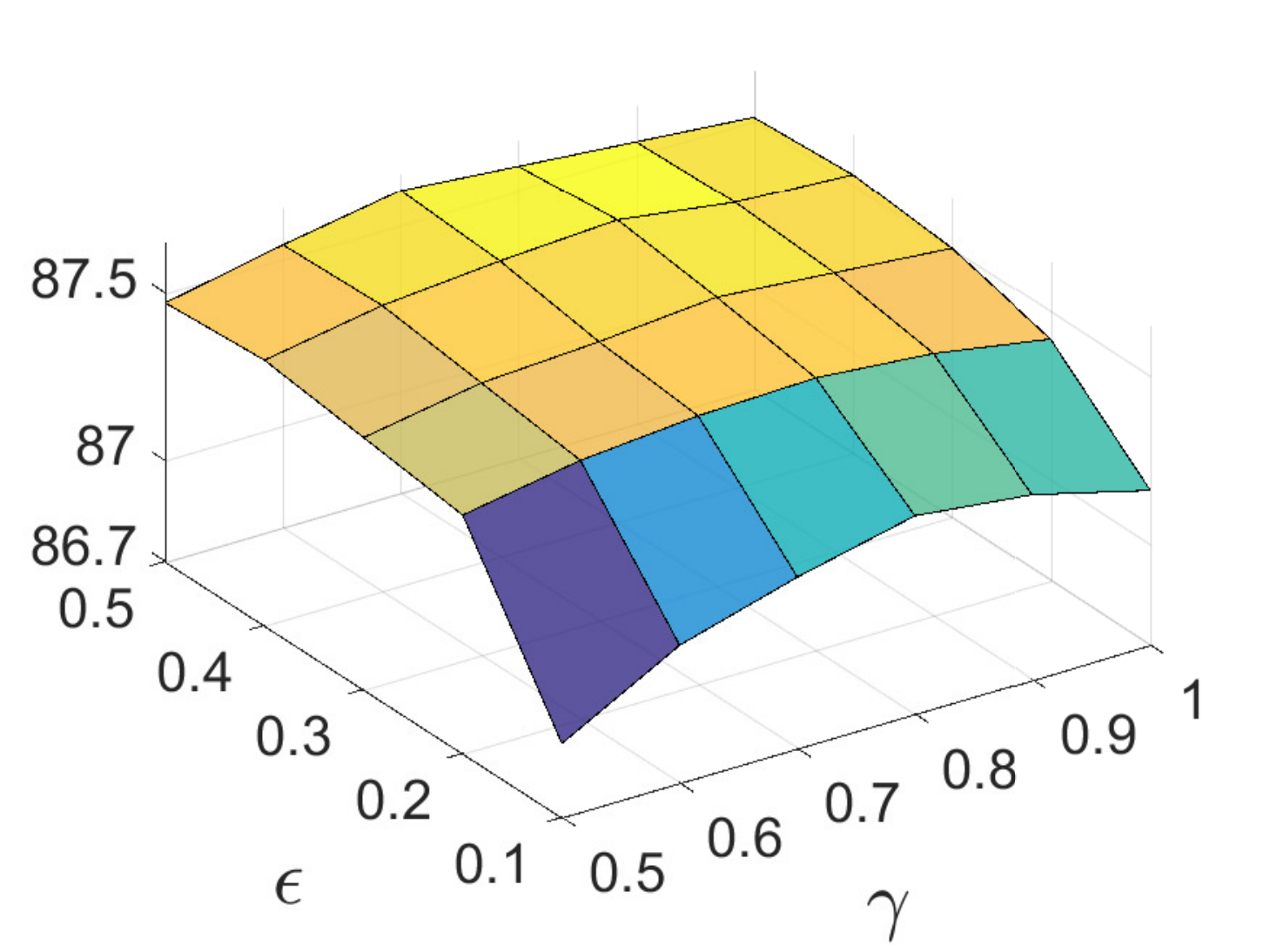}
		\caption{\texttt{Pubmed}}
	\end{subfigure}    \hfill
	\begin{subfigure}[t]{0.48\linewidth}
		\includegraphics[width=\textwidth]{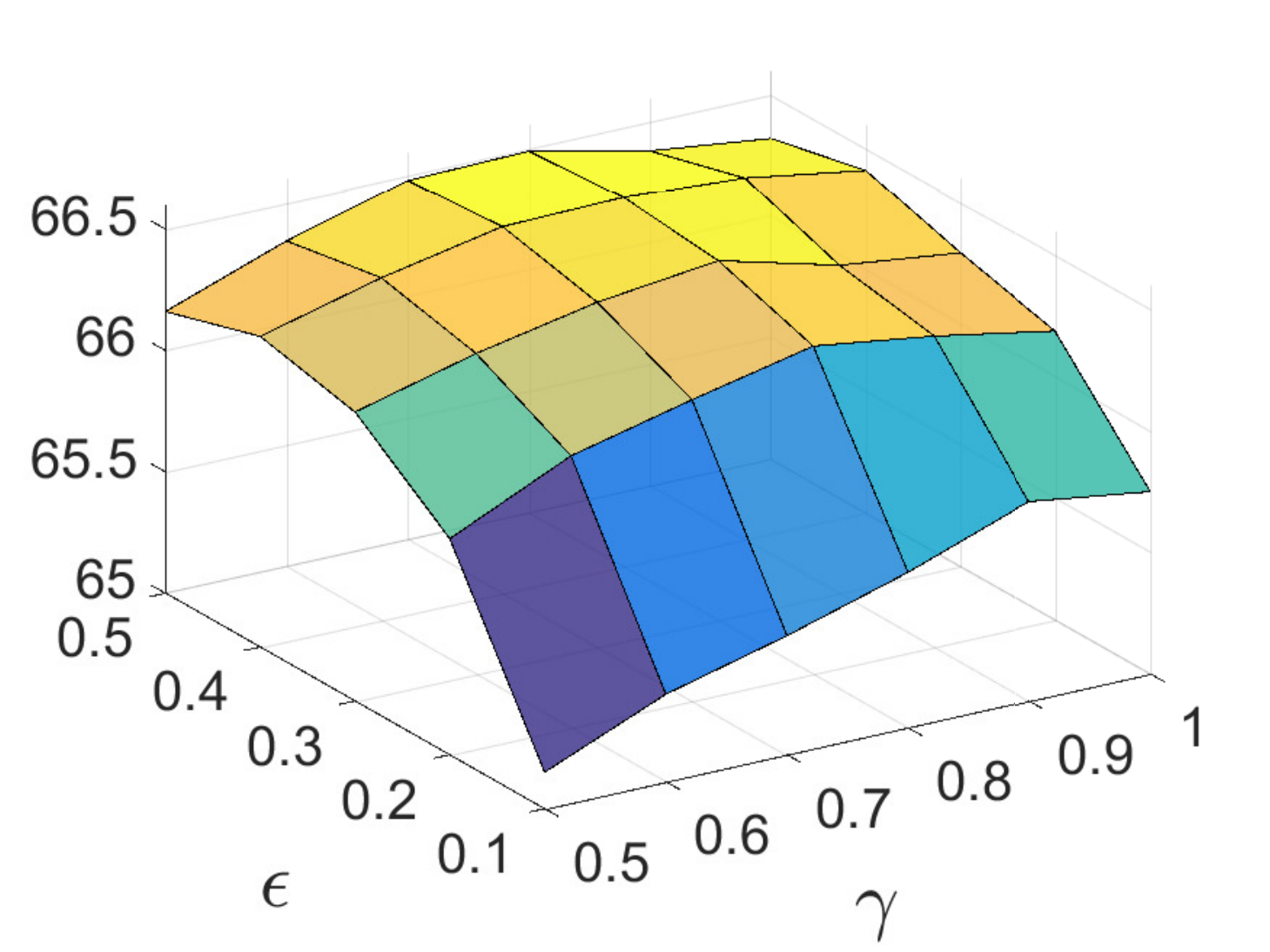}
		\caption{\texttt{CoraFull}}
	\end{subfigure}
	\caption{The test accuracy (z-axis) of GCN with SALS in (a) and ResGCN with SALS in (b) under different values of the hyper-parameters $\epsilon$ and $\gamma$.
		\label{fig:para}}
\end{figure}

Fig. \ref{fig:cali} shows the 10-bin reliability diagram of a 3-layer GCN trained on Cora \cite{guo2017calibration}. 
The grey dashed line represents perfect calibration where the output likelihood (confidence) perfectly predicts the accuracy.
Without label smoothing, the model trained with hard targets is clearly over-confident, since in expectation the accuracy is always below the confidence. 
With our SALS, we observe that the reliability diagram slope is now much closer to a slope of 1 and the model is better calibrated than the original GCN and it with LS. 
This meets our analysis in Sec. \ref{sec:3.2}, that the structure-aware soft targets correspond to suitable semantic positions that match the GNNs' inductive bias.
Our SALS smoothens the targets to make the labels of connected nodes correlated, which aligns better with the characteristics of graph data than the hard targets, LS, and OLS.

Last but not least, we evaluate how sensitive our SALS is to the hyper-parameters: $\epsilon$ that controls effects of label smoothing, and $\gamma$ that balances the neighborhood label and the pre-defined uniform distribution.
We visualize the results in Fig. \ref{fig:para}. 
The performance of GNNs with SALS is generally smooth when parameters are in certain ranges. 
However, too small values of $\epsilon$ and $\gamma$ result in low performances, which should be avoided in practice.
Moreover, increasing $\epsilon$ from 0.1 to 0.3 and $\gamma$ from 0.5 to 0.7 improves the effectiveness of SALS on all datasets, demonstrating that the label smoothing guided by graph structures plays an important role in improving the performance of GNNs.

\section{Conclusion}
In this work, we propose Structure-aware Label Smoothing (SALS) as an enhancement component to popular node classification models. 
SALS can capture the prior knowledge about the graph structures for supervising the label distributions. 
It produces structure-aware soft targets for every node in an adaptive manner.
Experiments on seven benchmark datasets proved our SALS's enhancement on various models for both transductive and inductive node classification.
Specifically, SALS reduces the over-fitting and calibration errors of GCN models, which uses the graph structural knowledge to enrich the label distribution
An interesting future direction is to extend our methods to other modalities such as text and images, where we can construct graph topology in the semantic space and extend our SALS to learn better label distributions.

\bibliographystyle{named}
\bibliography{ijcai22}

\end{document}